\pdfoutput=1

\documentclass[10pt,twocolumn,letterpaper]{article}

\usepackage[pagenumbers]{cvpr}

\makeatletter
\renewcommand\paragraph{\@startsection{paragraph}{4}{\z@}{1.5ex \@plus .5ex \@minus .2ex}{-1em}{\normalfont\normalsize\bfseries}}
\makeatother

\usepackage{multirow}
\usepackage{colortbl}
\definecolor{tabband}{HTML}{EEF1F5}
\definecolor{tabours}{HTML}{DDE8F6}
\definecolor{tabavg}{HTML}{E4E6EA}
\usepackage{makecell}
\usepackage{pifont}
\newcommand{\method}{\textsc{Heave-Ho}\xspace}
\newcommand{\bench}{\textsc{HH-Bench}\xspace}
\newcommand{\gptsix}{GPT-6 (Astra)\xspace}
\newif\ifarxiv
\newcommand{\projectpage}{project page}   %

\arxivtrue
\definecolor{cvprblue}{rgb}{0.21,0.49,0.74}
\usepackage[pagebackref,breaklinks,colorlinks,allcolors=cvprblue]{hyperref}
\usepackage{fontawesome5}
\renewcommand{\projectpage}{\href{https://heaveho-humanoid.github.io/}{project page}}
\AtEndPreamble{\crefname{appendix}{App.}{Apps.}\Crefname{appendix}{Appendix}{Appendices}\crefname{subappendix}{App.}{Apps.}\Crefname{subappendix}{Appendix}{Appendices}}

\newcommand{\projecturl}{\href{https://heaveho-humanoid.github.io/}{\faGlobe\ Project Page}}

\title{Heave-Ho: Decentralized Long-Horizon Multi-Humanoid Collaboration}

\author{
Ting-Hsuan Chen$^{1*}$ \quad Siheng Zhao$^{1*\ddagger}$ \quad Han Jiang$^{1*}$ \quad Qi Pan$^{1}$ \quad Louison Lu$^{1}$ \\
Jiaheng Luo$^{1}$ \quad Henghui Bao$^{1}$ \quad Yiyu Chen$^{2}$ \quad Quan Nguyen$^{1}$ \quad Yue Wang$^{1\dagger}$ \\[4pt]
$^{1}$University of Southern California \qquad $^{2}$VinMotion \\[2pt]
{\normalsize $^{*}$Equal contribution \quad$^{\dagger}$Corresponding author \quad$^{\ddagger}$Project lead} \\[2pt]
{\normalsize\projecturl}
}

\begin{document}
\twocolumn[{
\renewcommand\twocolumn[1][]{#1}
\maketitle
\vspace{-6mm}
\begin{center}
  \captionsetup{type=figure}
  \includegraphics[width=\textwidth]{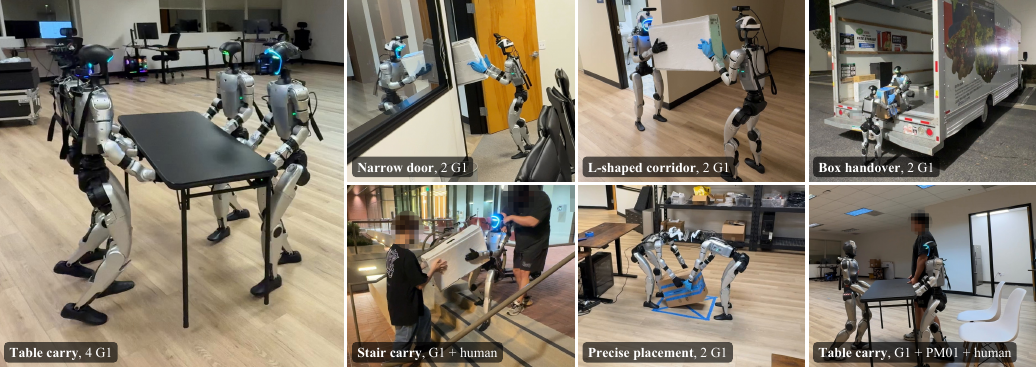}
  \captionof{figure}{\method enables decentralized humanoid teams to complete long-horizon collaborative tasks. Each robot acts on its own observations, the leader's planner chooses the next team skill and speaks it, and followers execute it in synchrony without any network communication. Videos can be found on our \href{https://heaveho-humanoid.github.io/}{project page}.}
  \label{fig:teaser}
\end{center}
}]
\begin{abstract}
Collaboration promises to extend humanoid capabilities beyond what individual robots can achieve.
However, realizing this potential requires coordinated whole-body interactions, coherent long-horizon planning, and precise synchronization across robots.
We introduce \method, a decentralized framework for long-horizon collaborative multi-humanoid loco-manipulation.
At the low level, we develop a multi-agent post-training framework that adapts a humanoid foundation tracking model to physically coupled interactions between humanoids, enabling coordinated decentralized control from local observations alone.
At the high level, we train a vision-language planner and design a speech-based leader–follower coordination mechanism to synchronize decentralized execution without network communication.
Trained with two identical humanoids, our system scales to teams of up to ten robots and generalizes to heterogeneous teams without retraining.
To benchmark these capabilities, we introduce \bench, comprising 12 tasks involving teams of 2–10 humanoids, including heterogeneous teams. 
Systematic evaluations show that \method consistently outperforms baseline methods, while real-world experiments demonstrate robust decentralized collaboration across challenging tasks.
\end{abstract}

\section{Introduction}
\label{sec:intro}

Learning-based whole-body control has made individual humanoids increasingly capable~\cite{he2025asap,zhao2025resmimic,luo2025sonic,ladderman}.
However, many tasks in homes, warehouses, and construction sites remain beyond the physical capabilities of a single robot.
Handling large or heavy objects, such as tables, bed frames, and large boxes, often requires multiple humanoids to share the load and coordinate their motion.
Collaboration therefore promises to overcome these physical limits.

Realizing this promise involves three interdependent challenges.
First, interaction is \emph{tightly coupled}: forces transmitted through a shared object make each robot's balance and grasp depend on its partners' actions.
Second, \emph{long-horizon} tasks require visually grounded planning of coordinated action sequences, such as reorienting the team and its shared load before passing through a narrow doorway.
Third, execution must be \emph{synchronized}: a mistimed lift or step can destabilize the shared load and the robots.

These challenges become especially demanding under decentralized execution.
A centralized policy can use team-wide observations to jointly coordinate the robots' actions and align their execution timing~\cite{michael2009cooperative,tuci2018cooperative,mandi2024roco,chen2024scalable}.
However, its input context and joint action space grow with team size, so policies with fixed input and output dimensions require modification and often retraining when robots are added.
Gathering team-wide observations also requires network communication at inference.
A shared decentralized policy can instead preserve the same per-robot observation and action interface as the team grows, but must coordinate from local observations alone.
Our goal is to combine this scalability with tightly synchronized whole-body collaboration, without network communication.

We introduce \method, a decentralized framework for long-horizon collaborative multi-humanoid loco-manipulation.
At the low level, our multi-agent post-training framework augments a frozen humanoid foundation tracking model~\cite{luo2025sonic} with a residual policy shared across robots.
We train this policy with multi-agent reinforcement learning using a centralized critic and shared actor, while each robot relies only on local observations at execution.
The resulting controller executes lifting, stepping, and turning skills with compatible start and end configurations, enabling their composition into long-horizon behaviors.
At the high level, we train a vision-language planner to select the next team skill from a leader robot's visual observations.
A speech-based leader--follower mechanism aligns skill decisions and execution timing without network communication: the leader announces a skill, followers recognize it locally, and all robots begin on a shared execution schedule.
The leader role can be assigned to any robot in the team, and the same speech interface allows a person to follow commands or act as the leader.

To evaluate these capabilities, we introduce \bench, a benchmark of 12 collaborative tasks for teams of 2--10 humanoids, including heterogeneous teams.
The tasks span lifting and carrying, shoulder carrying, pulling, pushing, handover, and precise placement under spatial constraints such as narrow doors and corridors.
Collaborative skills and task stages can be chained into long-horizon tasks.
Across these tasks, our low-level policy succeeds in 94\% of simulated episodes, compared with at most 29\% for baseline and ablated controllers.
Our planner completes 43 of 50 long-horizon trials, compared with 21 for zero-shot \gptsix{}, while reducing per-decision latency by a factor of about 120.
Without retraining, the policy learned with two identical humanoids scales to teams of up to ten robots and collaborates with a different humanoid platform and with people.
Extensive real-world experiments further demonstrate robust decentralized collaboration on long-horizon tasks.
Videos of the real-world and simulated experiments are provided on the \projectpage.

In summary, our contributions are:
\begin{itemize}
  \item A multi-agent post-training framework that adapts a frozen humanoid foundation tracking model to physically coupled collaboration through a shared residual policy, for decentralized control from local observations.
  \item A trained vision-language model for coherent long-horizon planning, together with a speech-based leader--follower coordination mechanism for synchronized decentralized execution without network communication.
  \item \bench, a benchmark of 12 collaborative tasks for teams of 2--10 humanoids, accompanied by systematic evaluations of scalability, heterogeneous collaboration, and real-world deployment.
\end{itemize}

\section{Related Work}
\label{sec:related}

\paragraph{Humanoid whole-body control and loco-manipulation.}
Learning-based methods have substantially advanced humanoid locomotion and athletic control~\cite{radosavovic2024humanoid,radosavovic2024challenging,zhuang2024parkour,wu2026php,ladderman,haarnoja2024soccer,kong2026learningsoccerskillshumanoid}.
To broaden these capabilities, motion imitation provides a scalable route to acquiring diverse whole-body skills from human demonstrations~\cite{cheng2024exbody,ji2024exbody2,he2024h2o,he2024omnih2o,he2025hover,he2025asap,fu2024humanplus,ze2025twist,ze2025twist2,liao2025beyondmimic,chen2025gmt}.
SONIC~\cite{luo2025sonic} further scales motion tracking into a generalist controller with a unified interface for teleoperation and general tracking.
Recent work extends motion tracking to contact-rich interaction using motion capture data~\cite{yang2025omniretarget,zhao2025resmimic,nguyen2026hotice} or human videos~\cite{weng2025hdmi,allshire2025videomimic,wang2026prism}.
Despite this progress, these methods primarily focus on the capabilities of individual humanoids.
Our work adapts a frozen foundation tracking model~\cite{luo2025sonic} through multi-agent residual learning, enabling physically coupled multi-humanoid loco-manipulation collaboration with decentralized control from local observations alone.

\paragraph{Centralized and decentralized multi-robot collaboration.}
Centralized policies jointly process team observations and predict actions for multiple robots, supporting multi-arm manipulation~\cite{tung2021learning,wang2025gaudp}, collaborative legged pushing~\cite{nachum2019multi}, and collaboration among heterogeneous robots~\cite{zhao2026duet}.
Decentralized approaches learn coordinated policies through multi-agent reinforcement learning with centralized training and decentralized execution~\cite{lowe2017maddpg,yu2022mappo} or joint imitation learning with diffusion policies and consensus representations~\cite{dong2025mimicd,he2025latent}.
CHORUS~\cite{doshi2026chorus} further extends this direction to a shared pretrained vision-language-action model, enabling heterogeneous mobile manipulators to act independently from their own observations.
Closely related to our setting, decPLM~\cite{pandit2026decplm} learns decentralized pinch--lift--move control for quadruped--arm teams, assuming externally specified contact locations, synchronized initiation, and payload motion commands.
In humanoid settings, existing work primarily investigates collaboration with human partners~\cite{bethala2025h2compact,du2026cola,kumbhar2025compliant,zhang2026c2c}.
Cooperative manipulation among multiple simulated characters has also been explored, which is confined to simulation only~\cite{gao2024coohoi,lionar2026teamhoi,yao2026synagent}.
Our work targets autonomous long-horizon collaborative loco-manipulation by real humanoid teams, combining decentralized whole-body control with visually grounded skill planning and synchronized execution without network communication.

\paragraph{Vision-language task planning.}
Language and vision-language models support long-horizon robot tasks by selecting and composing skills~\cite{huang2022zeroshot,ahn2022saycan,huang2022inner,driess2023palme} or generating executable control programs~\cite{liang2023cap}.
At the execution level, vision-language-action models map visual observations and language instructions to robot actions~\cite{brohan2023rt1,brohan2023rt2,kim2024openvla,black2024pi0,intelligence2025pi05,team2025gemini,bjorck2025gr00t,wei2026psi0}.
Hierarchical frameworks connect these levels by coupling task reasoning with learned visuomotor policies~\cite{shi2025hirobot} or humanoid motion and skill controllers~\cite{xue2025leverb,ding2025humanoidvla,yuan2025being0}.
For robot teams, centralized language planners allocate tasks using team-level information~\cite{kannan2024smartllm,liu2025coherent}, while decentralized and hybrid planning frameworks coordinate robot-specific agents through explicit inter-agent dialogue~\cite{mandi2024roco,chen2024scalable}, which need network communication.
In contrast, \method coordinates physically coupled humanoids without inter-robot network communication: a trained vision-language planner selects shared team skills from the leader's local visual observations, and speech-based coordination synchronizes their execution.
A shared low-level policy realizes each skill through decentralized whole-body control using only each robot's local observations.

\section{Method}
\label{sec:method}

\begin{figure*}[t]
  \centering
  \includegraphics[width=\linewidth]{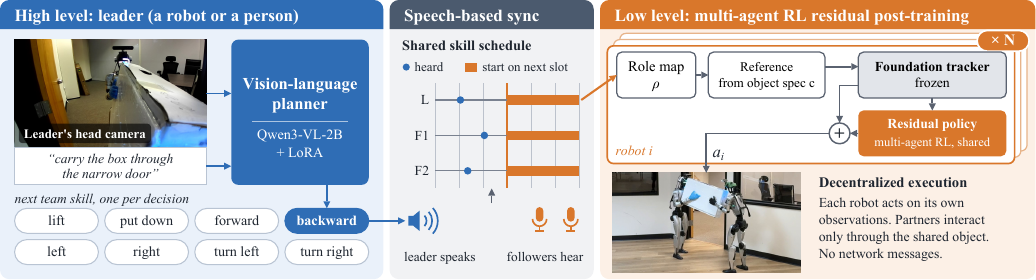}
  \caption{Overview of \method. At the high level, the leader's vision-language planner selects the next team skill from its camera and speaks it; every follower recognizes the command on board, and all robots start it on a shared skill schedule. At the low level, each robot maps the team skill to its own role, generates a reference conditioned on the object specification, and tracks it with a frozen foundation tracking model plus a residual policy shared by all robots. No network message is exchanged between robots.}
  \label{fig:overview}
\end{figure*}

\subsection{Overview}
\label{sec:formulation}
We consider a team of $N$ humanoids manipulating a shared object.
We call the team \emph{decentralized} in a strict sense: every robot acts only on its own observations, and no network message is exchanged between robots during a task.
The low-level policy runs on each robot's on-board computer.
Agent $i$ therefore observes only its proprioception, a short window of its own future reference motion, and the object specification $\mathbf{c}=(L,W,H,m)$ (length, width, height, and mass); it has no access to its partners' states, and their actions affect it only through forces transmitted by the shared object.

\method operates at two levels (\cref{fig:overview}).
The two levels communicate through a vocabulary of \emph{team skills} $\mathcal{K}$, such as lift, put down, a step in each direction, and a rotation in place.
The high level decides which team skill comes next, every few seconds, and makes sure that all robots start it together.
The low level turns the current team skill into whole-body motion on every robot at 50\,Hz.
Each robot maps a team skill to its own frame through a fixed role map $\rho_i: \mathcal{K}\to\mathcal{K}$; for two robots facing each other across a box, moving the team toward the leader's front is a forward step for the leader and a backward step for its partner, whereas lifting and rotating are the same for both.

\subsection{Low Level: Multi-Agent Post-Training}
\label{sec:low}

\paragraph{Composable team skills.}
Every team skill starts and ends at one of two static hub poses: $S_0$, standing empty-handed, and $S_1$, standing while holding the object.
Lift moves the team from $S_0$ to $S_1$, put down returns it to $S_0$, and steps and rotations keep it at $S_1$.
Because all skills share these endpoints, any skill can follow any other, and a long-horizon behavior reduces to a sequence of discrete choices.
Skills are defined at the team level: all robots start and end every skill in one canonical team layout and follow one common displacement profile during a move, so the spacing between partners stays constant over long sequences.
Steps have a fixed length and rotations a fixed angle, so the planner can reason about distance in units of skills, and when no new skill arrives the team simply holds its hub pose.

\paragraph{Reference motions.}
We obtain references from two sources.
We record pairs of people performing the task, recover their motion from video~\cite{shen2024gvhmr}, and retarget it to the robots; we also generate references procedurally with inverse kinematics, which adds objects and interaction modes without new recordings.
All references are cut into hub-to-hub skills and conditioned on $\mathbf{c}$: hand spread, grasp height, squat depth, and partner spacing adapt to the object, while the leg motion of each step is shared across objects.

\paragraph{Residual post-training.}
Learning balance and locomotion from scratch under multi-agent coupling is sample-inefficient and fragile.
We freeze a foundation tracking model~\cite{luo2025sonic} as a base policy $\pi_{\text{base}}$, which maps proprioception and reference to joint targets, and learn a residual policy $\pi_\theta$ that observes and corrects the base action:
\begin{equation}
  a_i = \pi_{\text{base}}(o_i) + \pi_\theta\big(o_i,\ \pi_{\text{base}}(o_i)\big).
  \label{eq:residual}
\end{equation}
The base policy provides balance and stepping, and the residual learns the forces and contacts needed to share a load with partners.
Exploration starts with a small action noise, so early rollouts stay close to the base policy.

\paragraph{Multi-agent reinforcement learning.}
All agents share the parameters $\theta$ of one actor~\cite{christianos2021sharing}, which keeps the policy independent of the team size.
We train with multi-agent PPO~\cite{schulman2017ppo,yu2022mappo} under centralized training and decentralized execution~\cite{lowe2017maddpg,foerster2018coma}: a centralized critic $V_\phi(s) \in \mathbb{R}^N$ receives the observations of all agents, concatenated in a fixed agent order, and outputs one value per agent, of which agent $i$ uses the $i$-th entry, while each actor acts only on its own observation.
Rewards are computed per agent and measure how well the agent tracks its own reference and how well the object follows its reference relative to the agent, with a contact term that rewards keeping the hands loaded while carrying and a gait term that rewards foot clearance and air time and penalizes dragging the feet.
Terminations are coupled: when any agent fails, the episode ends for the whole team, so all agents are trained under a shared outcome.

\paragraph{Training for composition and generalization.}
The skill is encoded only in the reference, so one network executes every skill and transitions never switch controllers.
To cover all transitions, we train on long random chains of legal skills with balanced directions and random waiting times at the hubs, and initialize episodes at random points of these chains~\cite{peng2018deepmimic}.
The object specification $\mathbf{c}$ is an explicit input; during training the true size varies by $\pm$10\% and the mass by $\pm$20\% around the nominal value given to the policy, so a single network adapts its grasp and posture to different objects.
We further randomize friction, robot properties, and external pushes, and apply curricula on actuation delay and on the object tracking tolerance~\cite{tobin2017domain,peng2018sim2real}.

\subsection{High Level: Planning and Synchronization}
\label{sec:high}

\begin{figure}[t]
  \centering
  \includegraphics[width=\linewidth]{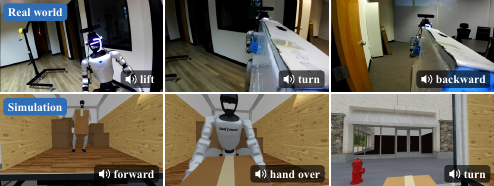}
  \caption{Leader's head-camera views over a task, each with the team skill chosen at that step: the narrow door in the real world (top) and the box handover in simulation (bottom).}
  \label{fig:ego}
  \vspace{-4mm}  %
\end{figure}

\paragraph{Vision-language planner.}
A leader decides the next team skill with a vision-language model.
Given the leader's head-camera image and the task instruction, the model outputs one token from $\mathcal{K}\cup\{\textsc{hold},\textsc{done}\}$.
It is queried after each skill completes, so every decision is made from a view of the team at rest, and a long-horizon task unfolds as a sequence of such decisions (\cref{fig:ego}).
We fine-tune Qwen3-VL-2B~\cite{bai2025qwen3vl} with LoRA~\cite{hu2022lora} (settings in App.~\ref{sec:supp_planner}) on episodes in which the leader's image is recorded whenever a team skill is committed.
In the real world, an operator selects the skills through the same interface, and we collect 24 episodes for each task of \cref{sec:exp_planner}; in simulation, a scripted planner selects them, and we collect 1,000 episodes for each task with initial positions sampled over a wide range.
A single decision maker gives the team one coherent plan, which independent planners on each robot cannot guarantee.
The planner can run on any computer that only the leader is connected to, since it never exchanges information with the other robots, and it is optional: without it, a person can act as the leader.

\paragraph{Speech-based leader-follower synchronization.}
Starting a skill together is critical: a late lift tilts the object, and a lone step pulls a partner off balance.
We separate \emph{what} the team does from \emph{when} it does it.
What is decided once, by the leader.
When is fixed by a shared skill schedule: a skill may only start on a coarse grid of control ticks, so robots that learn about the same skill at slightly different moments still start it together.
The decision is a single discrete token, which the leader speaks through its speaker; every follower recognizes it with its on-board microphone, and every robot schedules the skill on the next slot of the grid, with the leader offsetting its own start by the expected acoustic delay.
Because the grid is coarser than the variation in acoustic delay, this variation does not change the start slot, and the team starts every skill together without exchanging a single packet.
The leader is a role rather than a fixed robot, since every robot can host the same planner, and speech lets people join the team: a person can follow the spoken skills or take over as leader.

\subsection{Teams beyond Training}
\label{sec:scaling}
Nothing in the low-level policy depends on the number of partners: it is shared, its inputs are egocentric, and partners enter only through the object.
To deploy $N$ robots, we assign each robot a slot around the object, which defines its reference and contact targets, and run the same policy on all of them; four robots carrying a table, for instance, form two facing pairs side by side along its long edge.
Partners that do not run the G1 policy join through the same skill interface: a person follows the spoken skills, and a humanoid with a different body executes them with its own controller trained with the same recipe.

\vspace{-3mm}  %
\section{\bench: Multi-Humanoid Benchmark}
\label{sec:benchmark}

\begin{figure*}[t]
  \centering
  \includegraphics[width=\linewidth]{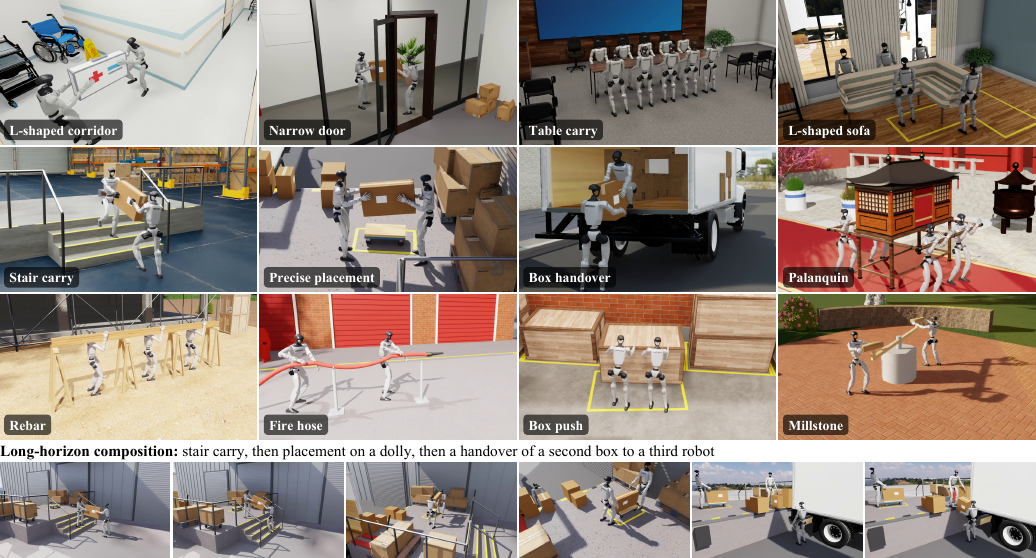}
  \caption{The tasks of \bench in simulation. Bottom: tasks can be chained into longer ones, here a stair carry, a placement on a dolly, and a handover of a second box to a third robot. Time sequences are in App.~\ref{sec:supp_sim} and videos on the \projectpage.}
  \label{fig:sim}
\end{figure*}

Evaluations of humanoid loco-manipulation mostly consider a single robot, and multi-agent benchmarks~\cite{samvelyan2019smac,carroll2019overcooked} rarely involve whole-body contact with a shared object.
There is thus no common testbed for humanoid teams that are physically coupled through a shared object, where the balance and grasp of each robot depend on its partners.
\bench collects 12 collaborative tasks in which a team of humanoids must act on the same object at the same time, often over long sequences of coordinated skills.
The tasks differ in how the robots interact with the object (lifting and carrying, carrying on the shoulder, pulling, pushing, handing over, and precise placement) and in how the environment constrains the motion (\cref{fig:sim}; App.~\ref{sec:supp_tasks}\ lists all tasks).
No task prescribes a team: each can be performed by teams of different sizes, including people and other humanoid platforms, in simulation or in the real world.
Together they test the three requirements of collaboration: tightly coupled whole-body interaction, coherent long-horizon planning, and synchronized execution.

\paragraph{Constrained transport.}
The narrow door is the signature task: the door is narrower than the object along its carrying direction, so the team first rotates the object in place and then walks through, with one robot moving forward and its partner backward.
A team that only walks along a fixed direction cannot solve it.
In the L-shaped corridor the team combines straight and sideways steps and must decide when to change direction while the robots and the object stay clear of the walls.
Carrying an L-shaped sofa onto a target area marked on the floor adds an irregular object whose load is distributed unevenly between the robots.
Precise placement requires the team to place a box within a target region, which calls for the richest set of gaits: walking forward, backward, and sideways, and turning in place.

\paragraph{Team scaling and heterogeneity.}
The table carry suits larger teams: a long table leaves room for many robots along its edges.
Any robot can be replaced by a person who follows the spoken skills, or by a humanoid with a different body; \cref{sec:exp_real,sec:exp_scaling} report the teams we evaluate.

\paragraph{Diverse interaction modes.}
Beyond lifting with the hands, the benchmark includes carrying a palanquin or a steel bar on the shoulders, pulling a fire hose, pushing a box or a millstone, and handing a box down from a robot on a higher level to a partner below, who must take over the load at a different height.
In our setup, the higher level is the bed of a truck, and the robot holding the box starts at different depths inside the truck and must first walk to its edge.
These tasks change where the contacts are and how force is shared, while all robots must still act together.

\paragraph{Composing long-horizon tasks.}
Because every task is executed as a sequence of hub-to-hub team skills, tasks can be chained into longer ones.
For example, two robots carry a box up a flight of stairs onto an elevated platform and place it on a wheeled dolly; one of them then leaves its partner, walks to the edge of the platform, and hands a second box of a different size down to a third robot standing below.
Such compositions test whether the team stays coordinated across stages that use different skills, and whether a robot can leave one team and join another within one task.

\paragraph{Inputs and evaluation.}
Robots use only on-board sensing and the nominal object specification; motion capture and network communication between robots are not allowed.
We report the success rate over 10 trials per task for real-world experiments and planners, and over three seeds of 512 teams for low-level controllers in simulation (\cref{sec:ablation}).
In simulation, the initial positions of the robots and the object are sampled over a wide range that covers both easy and hard configurations, whereas real-world trials start from a nominal configuration with small perturbations.
Success rates in simulation and in the real world are therefore not directly comparable.
A trial succeeds if the team completes the task and reaches the target location without dropping the object.

\section{Experiments}
\label{sec:experiments}

We evaluate \method on \bench and answer four questions.
(Q1) Can decentralized humanoid teams complete long-horizon collaborative tasks in the real world without communication?
(Q2) Does the high-level planner choose better team skills than alternative decision makers?
(Q3) How does the post-trained low-level policy compare with baseline controllers, and which of its design choices matter?
(Q4) Does a policy trained with two identical humanoids generalize to larger and heterogeneous teams?

\subsection{Setup}
\label{sec:setup}
We use Unitree G1 humanoids with a head-mounted stereo camera, and an EngineAI PM01 for heterogeneous teams.
The PM01 runs its own policy trained with the same recipe.
Policies run on each robot's on-board computer at 50\,Hz and are trained in Isaac-based simulation~\cite{makoviychuk2021isaacgym,mittal2025isaaclab} with 4,096 parallel two-robot environments~\cite{rudin2021legged}.

\subsection{Real-World Collaboration}
\label{sec:exp_real}
\begin{figure*}[t]
  \centering
  \includegraphics[width=\linewidth]{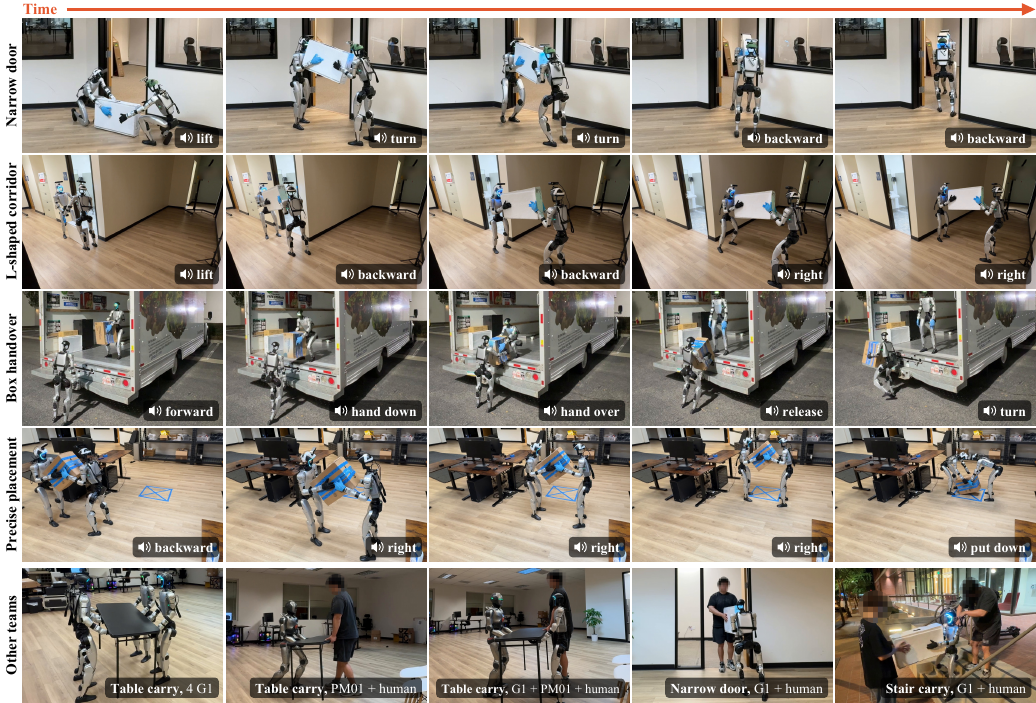}
  \caption{Real-world experiments. The first four rows show two G1 robots over time, from left to right, labeled with the team skill being executed. The last row shows teams that differ from the training configuration, including a different humanoid and people. Faces are obscured for anonymity. Full time sequences of all teams are in App.~\ref{sec:supp_real_seq} and videos on the \projectpage.}
  \label{fig:real}
\end{figure*}

\Cref{fig:real} shows real-world experiments of \method. Every robot runs the shared low-level policy on its own on-board computer, and no network message is exchanged between robots during a task.
Two-robot teams pass through the narrow door, follow the L-shaped corridor, hand a box down from a truck, and place a box within a target region.
A person can also take the place of one robot, carrying a box through the narrow door or up a flight of stairs together with a G1.
Finally, the two-robot policy is deployed to larger and heterogeneous teams that carry a table (\cref{sec:exp_scaling}).
A person on the team can also take over as the leader and speak the team skills, as we do in some of these experiments.
Every skill must start on all partners together, since a late lift tilts the object and a lone step pulls a partner off balance.
The videos on the \projectpage\ show the spoken commands and the synchronized start of each skill.
Across all tasks, the teams succeed in 92 of 100 trials (App.~\ref{sec:supp_real_results}\ lists every team).
Stair carry is the hardest (7/10): each step leaves room for little more than one foot, so small errors in foot placement are not tolerated.

\begin{figure*}[t]
  \centering
  \includegraphics[width=\linewidth]{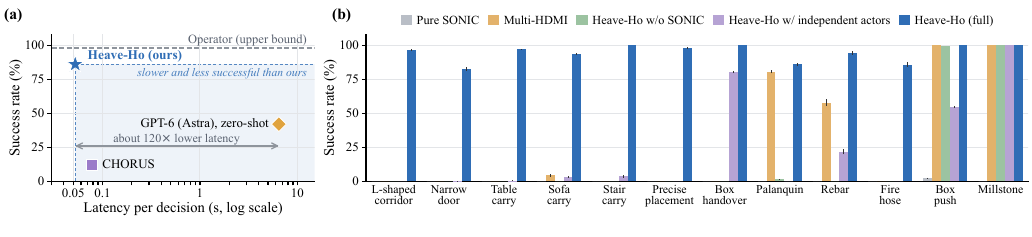}
  \caption{(a) Success rate of the decision makers, averaged over the five tasks of \cref{tab:planner}, against decision latency. The shaded region is slower and less successful than ours. The operator selects the skills by hand and serves as the upper bound. (b) Success rate of the low-level controllers on each task of \bench in simulation. Bars show the mean over three seeds, and error bars the standard deviation.}
  \label{fig:results}
\end{figure*}

\subsection{High-Level Planning}
\label{sec:exp_planner}
We evaluate the full system with different decision makers on five long-horizon tasks, with 10 trials per task (\cref{tab:planner} and \cref{fig:results}a).
Two run in the real world: the narrow door and the L-shaped corridor.
Three run in simulation: a box handover from the bed of a truck, in which the robot holding the box starts at different depths inside the truck and must walk to its edge before handing the box down; carrying a long table with four robots to the front of a sofa; and placing an L-shaped sofa on a target area marked on the floor.
We compare our fine-tuned Qwen3-VL-2B planner against two strong baselines: (1) zero-shot \gptsix{} with the same skill interface, and (2) CHORUS~\cite{doshi2026chorus}, a decentralized policy executed independently on each robot.
We fine-tune CHORUS on the same data following its recipe, including the $\pi_{0.5}$ backbone~\cite{intelligence2025pi05}, LoRA adapters, and a fixed identity prompt per robot, and change only its output: instead of action chunks, it selects team skills like our planner, which isolates the coordination problem.
An operator who selects the team skills manually serves as the upper bound.

Our planner matches the operator on all three simulated tasks and approaches it on the two real-world tasks, whereas \gptsix used zero-shot fails far more often on every task.
Our planner is also the fastest: one decision takes 52.5\,ms on a single RTX 4090 GPU with 5.3\,GiB of memory, compared with 78\,ms for CHORUS and 6.46\,s on average for \gptsix through its API, about 120 times longer.
CHORUS fails every trial on all tasks except the handover, where it succeeds in 6 of 10 trials.
Each robot runs its own copy of the policy and advances through its actions asynchronously, with no mechanism to align when the robots act~\cite{doshi2026chorus}.
The copies on different robots therefore start their skills at different moments, which breaks tasks in which the robots carry an object together.
The handover couples the robots only briefly: one robot carries the box to the truck's edge, and its partner approaches to receive it, which leaves CHORUS some chance of success.

\begin{table}[t]
  \centering
  \small
  \setlength{\tabcolsep}{3.5pt}
  \resizebox{\columnwidth}{!}{%
  \begin{tabular}{l cc ccc >{\columncolor{tabavg}}c}
    \toprule
    & \multicolumn{2}{c}{Real} & \multicolumn{3}{c}{Sim} & \\
    \cmidrule(lr){2-3} \cmidrule(lr){4-6}
    Decision maker & Door & Corridor & Handover & Table & Sofa & Avg. \\
    \midrule
    GPT-6 Astra (zero-shot)            & 3/10 & 4/10  & 7/10  & 4/10  & 3/10  & 42\% \\
    CHORUS~\cite{doshi2026chorus}  & 0/10 & 0/10  & 6/10  & 0/10  & 0/10  & 12\% \\
    \rowcolor{tabours}\method (ours)                 & \textbf{6/10} & \textbf{7/10} & \textbf{10/10} & \textbf{10/10} & \textbf{10/10} & \textbf{86\%} \\
    \midrule
    Human Operator         & 9/10 & 10/10 & 10/10 & 10/10 & 10/10 & 98\% \\
    \bottomrule
  \end{tabular}%
  }
  \caption{Success rate of the full system with different decision makers, 10 trials per task. Door: narrow door; Corridor: L-shaped corridor; Handover: box handover from a truck; Table: carrying a table to the front of a sofa with four robots; Sofa: placing an L-shaped sofa on a target area.}
  \label{tab:planner}
\end{table}

\subsection{Low-Level Post-Training}
\label{sec:ablation}
For each task, we compare the low-level policy against two baselines and two ablations in simulation, using the same skill chains, objects, and randomization (\cref{tab:ablation}).
\emph{Pure SONIC} tracks the same references with the frozen foundation tracker alone, without post-training, and is therefore unaware of the object and of its partners.
\emph{Multi-HDMI} extends the single-humanoid loco-manipulation method HDMI~\cite{weng2025hdmi} to the multi-humanoid setting, training it directly with independent PPO~\cite{dewitt2020ippo}, where each agent has its own actor and its own critic that sees only local observations.
\emph{w/o SONIC} removes the frozen foundation tracker from \method and learns the full action with the same multi-agent setup.
\emph{Independent actors} gives each agent its own actor instead of sharing parameters, while keeping the centralized critic.
\Cref{tab:ablation} and \cref{fig:results}b report the results.
\method succeeds in 94.41\% of the episodes on average, while no other controller exceeds 28.56\%.
Pure SONIC tracks the body reference more closely than Multi-HDMI and \method w/o SONIC but almost never completes a task: the foundation tracker reproduces the motion, yet does not know how to exert force on the object or share the load with its partners.
Without SONIC, learning the full whole-body action from scratch gives the largest body error and succeeds almost only on the two pushing tasks, which do not require lifting the object.
Starting from the frozen tracker instead, the residual only needs to refine tracking and interaction.
Independent actors reach 22.05\%.
With a shared actor, every update draws on the experience of all agents, which face similar but not identical situations in a collaborative task, and the number of parameters does not grow with the team size.
Multi-HDMI succeeds mainly on shoulder carry and pushing, and fails on tightly coupled lifting, where a critic that sees only local observations cannot tell how the partner contributes.

\begin{table}[t]
  \centering
  \small
  \setlength{\tabcolsep}{4pt}
  \resizebox{\columnwidth}{!}{%
  \begin{tabular}{lccc}
    \toprule
    Method & Success (\%) $\uparrow$ & Body (cm) $\downarrow$ & Object (cm) $\downarrow$ \\
    \midrule
    \rowcolor{tabband}\multicolumn{4}{l}{\emph{Baselines}} \\
    Pure SONIC~\cite{luo2025sonic} & 0.22 & 58.24 & 83.20 \\
    Multi-HDMI~\cite{weng2025hdmi} & 28.56 & 68.57 & 66.92 \\
    \rowcolor{tabband}\multicolumn{4}{l}{\emph{Ablations}} \\
    \method w/o SONIC & 16.78 & 92.69 & 89.53 \\
    \method w/ independent actors & 22.05 & 35.59 & 35.00 \\
    \midrule
    \rowcolor{tabours}\method (ours) & \textbf{94.41} & \textbf{18.62} & \textbf{19.59} \\
    \bottomrule
  \end{tabular}%
  }
  \caption{Low-level controllers in simulation, averaged over the 12 tasks of \bench, each with three seeds of 512 teams. Body and object errors are mean position errors over the full reference horizon, including frames after a failure. Per-task results are in App.~\ref{sec:supp_per_task}.}
  \label{tab:ablation}
  \vspace{-4mm}  %
\end{table}

\subsection{Train with Two, Deploy with Many}
\label{sec:exp_scaling}
\begin{figure}[t]
  \centering
  \includegraphics[width=\linewidth]{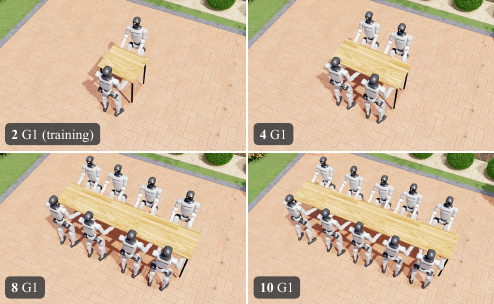}
  \caption{Train with two, deploy with many. The policy trained with two G1 robots (top left) carries a table in simulation with teams of four, eight, and ten robots, without retraining.}
  \label{fig:teamsize}
  \vspace{-4mm}  %
\end{figure}

All G1 policies are trained with exactly two identical robots.
We deploy the same policy, without retraining, to teams that differ from this configuration in size and in composition.
In the real world (\cref{fig:real}), four G1 robots carry a table as two pairs side by side, three G1 robots carry it with a person, a G1 carries it with a person and a PM01, and a PM01 carries it with a person; these teams succeed in 37 of 40 trials.
In simulation, the same policy drives teams of four, eight, and ten G1 robots carrying a table (\cref{fig:teamsize}).
Two properties make this possible: the shared actor has no notion of team size, and partners affect each robot only through the object.
Parameter sharing is essential, since independent actors would require a new actor for every additional robot.

\section{Conclusion}
\label{sec:conclusion}
We introduced \method, a decentralized framework for long-horizon collaborative multi-humanoid loco-manipulation.
At the low level, multi-agent post-training turns a frozen humanoid foundation tracking model into a shared controller for physically coupled whole-body interaction.
At the high level, a vision-language planner gives the team one coherent long-horizon plan, more accurately and about 120 times faster than \gptsix used zero-shot, and a speech-based leader-follower mechanism executes it in synchrony without network communication.
On \bench, \method completes long-horizon collaborative tasks in the real world, and a policy trained with two identical humanoids carries over to teams of up to ten robots, to a different humanoid platform, and to people.

\paragraph{Limitations.}
Synchronization relies on followers hearing the leader, so it requires a reasonably quiet environment and reliable sound pickup.
Training the planner for a new task requires demonstrations collected manually by teleoperation, which is time-consuming.

\section*{Acknowledgments}
We thank EngineAI for providing the PM01 humanoid robot.
Siheng Zhao was supported by an Amazon ML PhD Fellowship from the USC + Amazon Center on Secure \& Trusted Machine Learning.

{
    \small
    \bibliographystyle{ieeenat_fullname}
    \bibliography{main}
}

\clearpage
\appendix
\begin{center}
  {\Large\bf Supplementary Material}
\end{center}
\section{Human Participants}
\label{sec:supp_human}
The people who appear in our real-world experiments, figures, and videos are members of the research team and are among the authors.
They took part as carrying partners or as the leader who speaks the team commands.
No data about the participants themselves was collected or analyzed, and the evaluation measures only task success; therefore, no IRB review was sought.
All participants consented to appear in the figures and videos, and their faces are obscured to preserve anonymity.

\section{Benchmark Details}
\label{sec:supp_bench}

\subsection{Tasks}
\label{sec:supp_tasks}
\Cref{tab:benchmark} lists the 12 tasks of \bench, grouped by how the robots interact with the object; \cref{fig:sim}\ shows each of them in simulation.

\begin{table}[h]
  \centering
  \small
  \setlength{\tabcolsep}{5pt}
  \begin{tabular}{l l}
    \toprule
    Task & What it tests \\
    \midrule
    \rowcolor{tabband}\multicolumn{2}{l}{\emph{Lift and carry}} \\
    L-shaped corridor   & steps along a turning path \\
    Narrow door         & rotating to fit through a door \\
    Table carry         & a large object shared by many robots \\
    L-shaped sofa       & an irregular, unevenly loaded object \\
    Stair carry         & partners at different heights \\
    Precise placement   & placing a box in a target region \\
    \rowcolor{tabband}\multicolumn{2}{l}{\emph{Handover}} \\
    Box handover        & passing a load down to a partner \\
    \rowcolor{tabband}\multicolumn{2}{l}{\emph{Shoulder carry}} \\
    Palanquin           & a load carried above the arms \\
    Rebar               & a long load along a line of robots \\
    \rowcolor{tabband}\multicolumn{2}{l}{\emph{Pull}} \\
    Fire hose           & pulling a deformable object \\
    \rowcolor{tabband}\multicolumn{2}{l}{\emph{Push}} \\
    Box push            & pushing a heavy object together \\
    Millstone           & pushing around a fixed axis \\
    \bottomrule
  \end{tabular}
  \caption{Tasks in \bench, grouped by how the robots interact with the object.}
  \label{tab:benchmark}
\end{table}

\subsection{Simulated Time Sequences}
\label{sec:supp_sim}
\begin{figure*}[p]
  \centering
  \includegraphics[width=\textwidth]{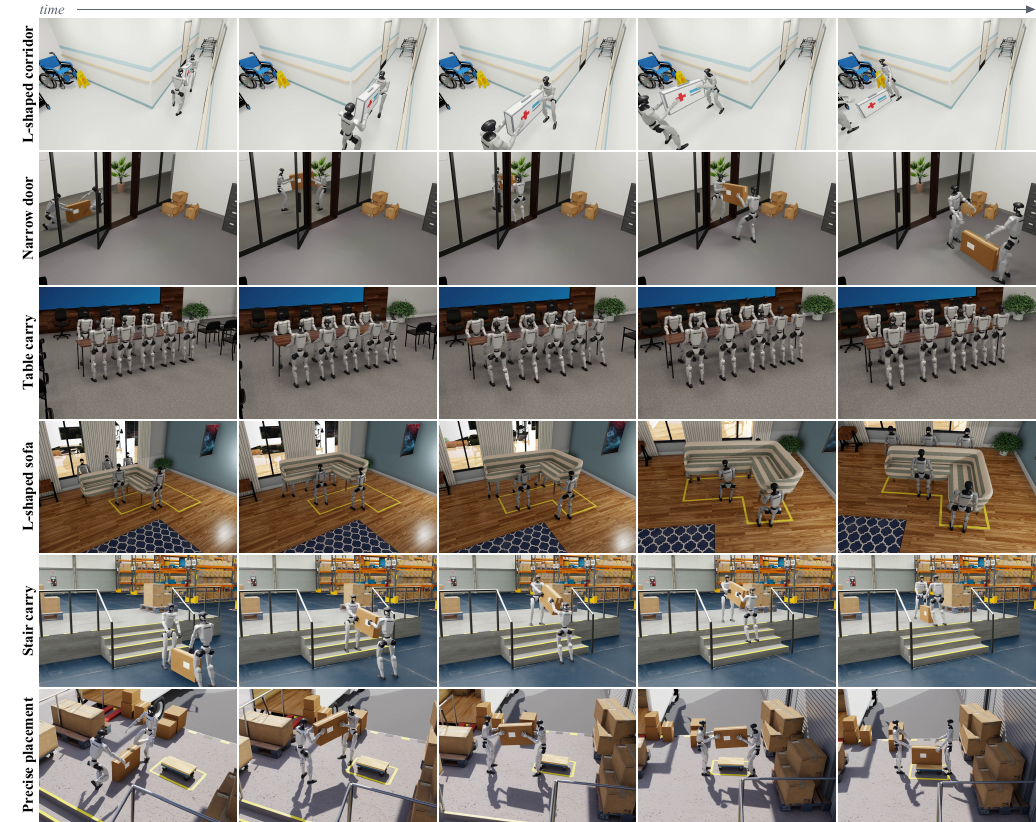}
  \caption{Time sequences of the lift-and-carry tasks in \bench, in simulation. Time runs from left to right.}
  \label{fig:supp_sim_a}
\end{figure*}

\begin{figure*}[p]
  \centering
  \includegraphics[width=\textwidth]{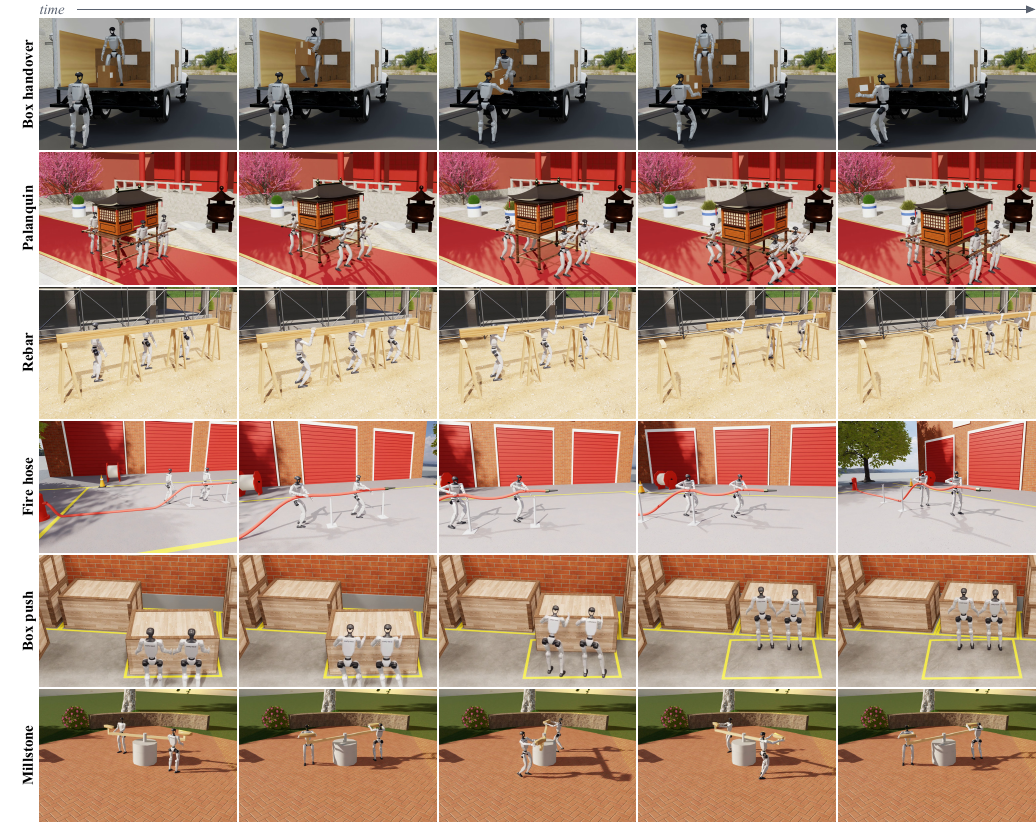}
  \caption{Time sequences of the handover, shoulder-carry, pull, and push tasks in \bench, in simulation. Time runs from left to right.}
  \label{fig:supp_sim_b}
\end{figure*}

\Cref{fig:supp_sim_a,fig:supp_sim_b} show a time sequence for each of the 12 tasks, in the order of \cref{tab:benchmark}.

\section{Real-World Experiments}
\label{sec:supp_realexp}

\subsection{Results per Team}
\label{sec:supp_real_results}
\Cref{tab:real} lists the success rate of every real-world team in \cref{sec:exp_real}.

\begin{table}[h]
  \centering
  \small
  \setlength{\tabcolsep}{4pt}
  \begin{tabular}{l l c}
    \toprule
    Task & Team & Success \\
    \midrule
    Narrow door         & 2 G1                 & 9/10 \\
    L-shaped corridor   & 2 G1                 & 10/10 \\
    Box handover        & 2 G1                 & 10/10 \\
    Precise placement   & 2 G1                 & 9/10 \\
    \midrule
    Narrow door         & G1 + human           & 10/10 \\
    Stair carry         & G1 + human           & 7/10 \\
    \midrule
    Table carry         & 4 G1                 & 8/10 \\
    Table carry         & 3 G1 + human         & 9/10 \\
    Table carry         & PM01 + human         & 10/10 \\
    Table carry         & G1 + PM01 + human    & 10/10 \\
    \bottomrule
  \end{tabular}
  \caption{Real-world experiments. No network message is exchanged between robots.}
  \label{tab:real}
\end{table}

\subsection{Time Sequences}
\label{sec:supp_real_seq}
\begin{figure*}[p]
  \centering
  \includegraphics[width=\textwidth]{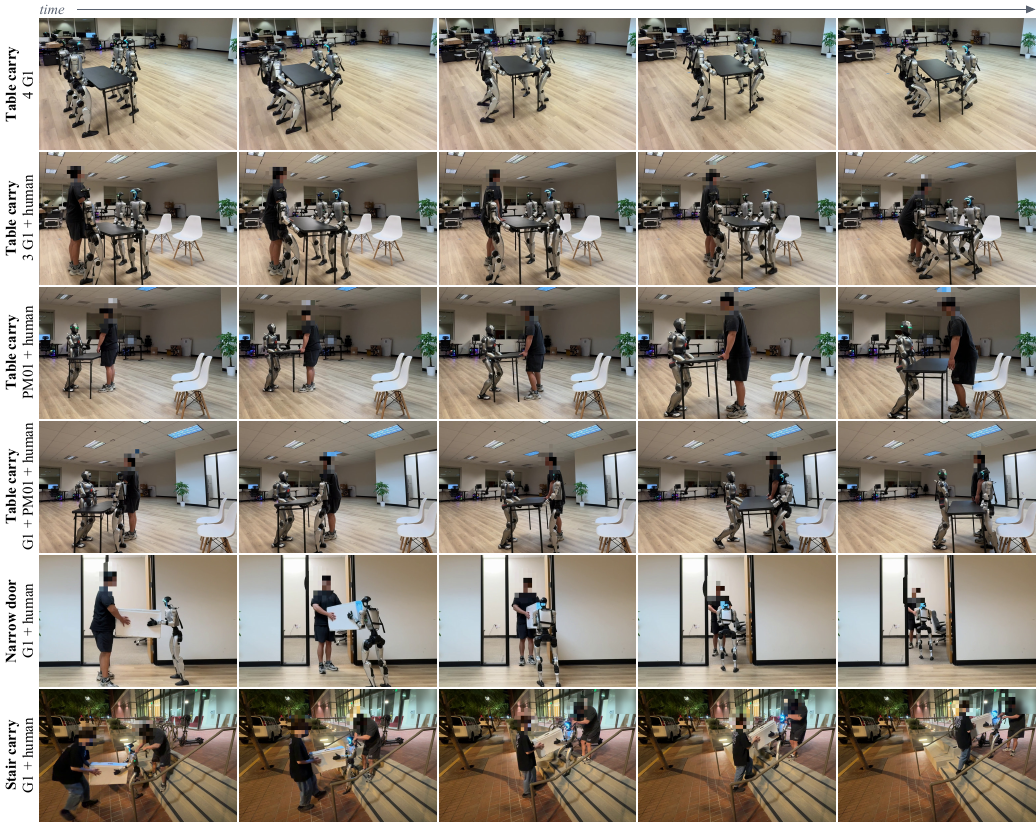}
  \caption{Real-world time sequences of the teams shown as single frames in \cref{fig:real}, and of the team of three G1 robots and a person. Time runs from left to right. Faces are obscured for anonymity.}
  \label{fig:supp_real}
\end{figure*}

\Cref{fig:real}\ shows time sequences for the two-robot teams and a single frame for each of the other teams.
\Cref{fig:supp_real} shows the full time sequences of these other teams, together with the team of three G1 robots and a person, which is listed in \cref{tab:real} but not shown in \cref{fig:real}.

\section{Implementation Details}
\label{sec:supp_impl}

\subsection{Planner Training}
\label{sec:supp_planner}
\Cref{tab:supp_planner} lists the training settings of the vision-language planner described in \cref{sec:high}.

\begin{table}[h]
  \centering
  \small
  \setlength{\tabcolsep}{4pt}
  \begin{tabular}{@{}l p{0.62\columnwidth}@{}}
    \toprule
    Item & Value \\
    \midrule
    Base model & Qwen3-VL-2B-Instruct \\
    LoRA rank / alpha & 32 / 64 \\
    LoRA targets (LLM) & q, k, v, o, gate, up, and down projections in every layer \\
    LoRA targets (vision) & attention qkv and output projections and both MLP layers in the last 8 ViT blocks \\
    Vision merger & fully trained \\
    Trainable parameters & $\approx$140\,M (4.2\,M in the ViT LoRA) \\
    Optimizer & AdamW, 5\% warmup, cosine decay \\
    Learning rate & $2\times10^{-4}$ (LLM), $5\times10^{-5}$ (ViT) \\
    Batch size & 4 $\times$ 4 gradient accumulation (effective 16) \\
    Epochs & 2 \\
    Precision & bf16 base weights, fp32 trainable parameters, gradient checkpointing \\
    Training data & 24 real-world episodes or 1,000 simulated episodes per task \\
    Inference & 52.5\,ms per decision, 5.3\,GiB, one RTX 4090 GPU \\
    \bottomrule
  \end{tabular}
  \caption{Training settings of the vision-language planner.}
  \label{tab:supp_planner}
\end{table}

\subsection{Low-Level Policy Training}
\label{sec:supp_low_level}
We post-train the low-level policy in Isaac Lab with teams of two Unitree G1 robots.
A shared actor predicts joint-target residuals on top of frozen SONIC, and a centralized critic sees the concatenated observations of both robots during training.
We train a teacher with privileged simulator observations and distill its mean actions into a student while training an adaptation module to predict the privileged latent from local inputs.
We then fine-tune the student with MAPPO in closed loop.
Both teacher and student use the same frozen SONIC base.
\Cref{tab:supp_low_level} gives the MAPPO settings.
For the box-carrying skills (lifting, putting down, stepping in each direction, and turning in place), which together cover the two-robot box-carrying tasks such as the narrow door and the L-shaped corridor, and for the box handover, training uses 4,096 parallel teams and a budget of $10^9$ environment steps per training stage, counting one step per team.

\begin{table}[h]
  \centering
  \small
  \setlength{\tabcolsep}{4pt}
  \begin{tabular}{@{}l p{0.50\columnwidth}@{}}
    \toprule
    Item & Value \\
    \midrule
    Physics / control time step & 5 / 20\,ms \\
    Actor hidden widths & 512, 256, 256 \\
    Critic hidden widths & 512, 256, 128 \\
    Optimizer & Adam \\
    Initial learning rate & $3\times10^{-4}$ \\
    Actor KL target & 0.01 \\
    Rollout length & 32 control steps per team \\
    PPO epochs / minibatches & 3 epochs, 8 minibatches per epoch \\
    PPO clip range & 0.2 \\
    Discount / GAE $\lambda$ & 0.99 / 0.95 \\
    Entropy coefficient & 0.001 \\
    Gradient norm limit & 1.0 \\
    \bottomrule
  \end{tabular}
  \caption{MAPPO settings for low-level policy post-training.}
  \label{tab:supp_low_level}
\end{table}

\section{Additional Results}
\label{sec:supp_results}

\subsection{Per-Task Low-Level Results}
\label{sec:supp_per_task}
\Cref{sec:ablation}\ compares the low-level controllers averaged over the 12 tasks of \bench (\cref{tab:ablation}) and shows their success rate on each task (\cref{fig:results}b).
\Cref{tab:supp_per_task_ablation} lists the success rate, body error, and object error of every controller on each task.

\begin{table*}[t]
  \centering
  \small
  \setlength{\tabcolsep}{3pt}
  \begin{tabular*}{\textwidth}{@{\extracolsep{\fill}}l *{5}{c}@{}}
    \toprule
    \textbf{Success rate (\%) $\uparrow$} & \textbf{Pure SONIC} & \textbf{Multi-HDMI} & \textbf{\method\ w/o SONIC} & \textbf{\method\ w/ indep. actors} & \textbf{\method} \\
    \midrule
    L-shaped corridor & 0.00\textsuperscript{\ensuremath{\pm}0.00} & 0.00\textsuperscript{\ensuremath{\pm}0.00} & 0.00\textsuperscript{\ensuremath{\pm}0.00} & 0.00\textsuperscript{\ensuremath{\pm}0.00} & \textbf{96.29}\textsuperscript{\ensuremath{\pm}0.68} \\
    Narrow door & 0.00\textsuperscript{\ensuremath{\pm}0.00} & 0.00\textsuperscript{\ensuremath{\pm}0.00} & 0.00\textsuperscript{\ensuremath{\pm}0.00} & 0.39\textsuperscript{\ensuremath{\pm}0.20} & \textbf{82.62}\textsuperscript{\ensuremath{\pm}1.55} \\
    Table carry & 0.00\textsuperscript{\ensuremath{\pm}0.00} & 0.00\textsuperscript{\ensuremath{\pm}0.00} & 0.00\textsuperscript{\ensuremath{\pm}0.00} & 0.52\textsuperscript{\ensuremath{\pm}0.30} & \textbf{97.01}\textsuperscript{\ensuremath{\pm}0.49} \\
    Sofa carry & 0.00\textsuperscript{\ensuremath{\pm}0.00} & 4.36\textsuperscript{\ensuremath{\pm}0.88} & 0.00\textsuperscript{\ensuremath{\pm}0.00} & 3.26\textsuperscript{\ensuremath{\pm}0.69} & \textbf{93.36}\textsuperscript{\ensuremath{\pm}0.70} \\
    Stair carry & 0.00\textsuperscript{\ensuremath{\pm}0.00} & 0.00\textsuperscript{\ensuremath{\pm}0.00} & 0.00\textsuperscript{\ensuremath{\pm}0.00} & 3.71\textsuperscript{\ensuremath{\pm}1.09} & \textbf{100.00}\textsuperscript{\ensuremath{\pm}0.00} \\
    Precise placement & 0.00\textsuperscript{\ensuremath{\pm}0.00} & 0.00\textsuperscript{\ensuremath{\pm}0.00} & 0.00\textsuperscript{\ensuremath{\pm}0.00} & 0.00\textsuperscript{\ensuremath{\pm}0.00} & \textbf{97.66}\textsuperscript{\ensuremath{\pm}0.70} \\
    Box handover & 0.00\textsuperscript{\ensuremath{\pm}0.00} & 0.00\textsuperscript{\ensuremath{\pm}0.00} & 0.00\textsuperscript{\ensuremath{\pm}0.00} & 80.40\textsuperscript{\ensuremath{\pm}0.98} & \textbf{100.00}\textsuperscript{\ensuremath{\pm}0.00} \\
    Palanquin & 0.00\textsuperscript{\ensuremath{\pm}0.00} & 80.53\textsuperscript{\ensuremath{\pm}1.24} & 1.56\textsuperscript{\ensuremath{\pm}0.34} & 0.00\textsuperscript{\ensuremath{\pm}0.00} & \textbf{85.87}\textsuperscript{\ensuremath{\pm}1.33} \\
    Rebar & 0.00\textsuperscript{\ensuremath{\pm}0.00} & 57.88\textsuperscript{\ensuremath{\pm}2.44} & 0.00\textsuperscript{\ensuremath{\pm}0.00} & 21.61\textsuperscript{\ensuremath{\pm}1.77} & \textbf{94.34}\textsuperscript{\ensuremath{\pm}1.19} \\
    Fire hose & 0.00\textsuperscript{\ensuremath{\pm}0.00} & 0.00\textsuperscript{\ensuremath{\pm}0.00} & 0.00\textsuperscript{\ensuremath{\pm}0.00} & 0.00\textsuperscript{\ensuremath{\pm}0.00} & \textbf{85.81}\textsuperscript{\ensuremath{\pm}1.58} \\
    Box push & 2.67\textsuperscript{\ensuremath{\pm}0.11} & \textbf{100.00}\textsuperscript{\ensuremath{\pm}0.00} & 99.80\textsuperscript{\ensuremath{\pm}0.00} & 54.75\textsuperscript{\ensuremath{\pm}0.56} & \textbf{100.00}\textsuperscript{\ensuremath{\pm}0.00} \\
    Millstone & 0.00\textsuperscript{\ensuremath{\pm}0.00} & \textbf{100.00}\textsuperscript{\ensuremath{\pm}0.00} & \textbf{100.00}\textsuperscript{\ensuremath{\pm}0.00} & \textbf{100.00}\textsuperscript{\ensuremath{\pm}0.00} & \textbf{100.00}\textsuperscript{\ensuremath{\pm}0.00} \\
    \bottomrule
  \end{tabular*}
  \par\smallskip
  \begin{tabular*}{\textwidth}{@{\extracolsep{\fill}}l *{5}{c}@{}}
    \toprule
    \textbf{Body error (cm) $\downarrow$} & \textbf{Pure SONIC} & \textbf{Multi-HDMI} & \textbf{\method\ w/o SONIC} & \textbf{\method\ w/ indep. actors} & \textbf{\method} \\
    \midrule
    L-shaped corridor & 89.24\textsuperscript{\ensuremath{\pm}0.14} & 227.81\textsuperscript{\ensuremath{\pm}0.80} & 269.41\textsuperscript{\ensuremath{\pm}0.98} & 64.98\textsuperscript{\ensuremath{\pm}0.58} & \textbf{39.51}\textsuperscript{\ensuremath{\pm}0.04} \\
    Narrow door & 84.59\textsuperscript{\ensuremath{\pm}0.11} & 149.78\textsuperscript{\ensuremath{\pm}0.63} & 182.96\textsuperscript{\ensuremath{\pm}0.46} & 41.32\textsuperscript{\ensuremath{\pm}2.36} & \textbf{16.71}\textsuperscript{\ensuremath{\pm}0.30} \\
    Table carry & 36.03\textsuperscript{\ensuremath{\pm}0.13} & 46.52\textsuperscript{\ensuremath{\pm}0.20} & 51.86\textsuperscript{\ensuremath{\pm}0.39} & 102.72\textsuperscript{\ensuremath{\pm}4.80} & \textbf{25.40}\textsuperscript{\ensuremath{\pm}0.37} \\
    Sofa carry & 31.81\textsuperscript{\ensuremath{\pm}0.14} & 92.22\textsuperscript{\ensuremath{\pm}1.57} & 95.67\textsuperscript{\ensuremath{\pm}0.93} & 73.72\textsuperscript{\ensuremath{\pm}0.69} & \textbf{25.63}\textsuperscript{\ensuremath{\pm}0.07} \\
    Stair carry & 106.64\textsuperscript{\ensuremath{\pm}0.42} & 54.23\textsuperscript{\ensuremath{\pm}0.29} & 115.92\textsuperscript{\ensuremath{\pm}0.83} & 13.05\textsuperscript{\ensuremath{\pm}0.30} & \textbf{11.04}\textsuperscript{\ensuremath{\pm}0.01} \\
    Precise placement & 67.15\textsuperscript{\ensuremath{\pm}0.07} & 120.49\textsuperscript{\ensuremath{\pm}0.19} & 142.76\textsuperscript{\ensuremath{\pm}0.64} & 41.02\textsuperscript{\ensuremath{\pm}1.41} & \textbf{16.11}\textsuperscript{\ensuremath{\pm}0.54} \\
    Box handover & 11.57\textsuperscript{\ensuremath{\pm}0.51} & 36.86\textsuperscript{\ensuremath{\pm}0.09} & 37.99\textsuperscript{\ensuremath{\pm}0.65} & 6.95\textsuperscript{\ensuremath{\pm}0.09} & \textbf{6.21}\textsuperscript{\ensuremath{\pm}0.01} \\
    Palanquin & 56.42\textsuperscript{\ensuremath{\pm}0.10} & 20.03\textsuperscript{\ensuremath{\pm}0.09} & \textbf{13.50}\textsuperscript{\ensuremath{\pm}0.40} & 22.52\textsuperscript{\ensuremath{\pm}0.16} & 22.14\textsuperscript{\ensuremath{\pm}0.12} \\
    Rebar & 36.21\textsuperscript{\ensuremath{\pm}0.02} & 22.35\textsuperscript{\ensuremath{\pm}0.43} & 50.17\textsuperscript{\ensuremath{\pm}0.35} & \textbf{15.03}\textsuperscript{\ensuremath{\pm}0.54} & 16.25\textsuperscript{\ensuremath{\pm}0.21} \\
    Fire hose & 40.22\textsuperscript{\ensuremath{\pm}0.09} & \textbf{22.83}\textsuperscript{\ensuremath{\pm}0.27} & 138.88\textsuperscript{\ensuremath{\pm}0.29} & 23.47\textsuperscript{\ensuremath{\pm}0.10} & 30.70\textsuperscript{\ensuremath{\pm}0.18} \\
    Box push & 38.71\textsuperscript{\ensuremath{\pm}0.35} & 9.45\textsuperscript{\ensuremath{\pm}0.03} & 8.93\textsuperscript{\ensuremath{\pm}0.01} & 15.22\textsuperscript{\ensuremath{\pm}0.04} & \textbf{7.65}\textsuperscript{\ensuremath{\pm}0.01} \\
    Millstone & 100.29\textsuperscript{\ensuremath{\pm}0.10} & 20.27\textsuperscript{\ensuremath{\pm}0.05} & \textbf{4.27}\textsuperscript{\ensuremath{\pm}0.01} & 7.10\textsuperscript{\ensuremath{\pm}0.04} & 6.04\textsuperscript{\ensuremath{\pm}0.00} \\
    \bottomrule
  \end{tabular*}
  \par\smallskip
  \begin{tabular*}{\textwidth}{@{\extracolsep{\fill}}l *{5}{c}@{}}
    \toprule
    \textbf{Object error (cm) $\downarrow$} & \textbf{Pure SONIC} & \textbf{Multi-HDMI} & \textbf{\method\ w/o SONIC} & \textbf{\method\ w/ indep. actors} & \textbf{\method} \\
    \midrule
    L-shaped corridor & 166.87\textsuperscript{\ensuremath{\pm}1.62} & 242.21\textsuperscript{\ensuremath{\pm}0.60} & 238.43\textsuperscript{\ensuremath{\pm}2.36} & 67.01\textsuperscript{\ensuremath{\pm}0.29} & \textbf{41.00}\textsuperscript{\ensuremath{\pm}0.03} \\
    Narrow door & 101.06\textsuperscript{\ensuremath{\pm}0.48} & 139.67\textsuperscript{\ensuremath{\pm}0.30} & 154.70\textsuperscript{\ensuremath{\pm}1.08} & 41.07\textsuperscript{\ensuremath{\pm}1.75} & \textbf{26.50}\textsuperscript{\ensuremath{\pm}0.34} \\
    Table carry & 52.51\textsuperscript{\ensuremath{\pm}0.84} & 44.50\textsuperscript{\ensuremath{\pm}0.70} & 51.09\textsuperscript{\ensuremath{\pm}0.69} & 83.84\textsuperscript{\ensuremath{\pm}11.16} & \textbf{24.21}\textsuperscript{\ensuremath{\pm}0.28} \\
    Sofa carry & 93.89\textsuperscript{\ensuremath{\pm}0.17} & 109.61\textsuperscript{\ensuremath{\pm}1.41} & 111.37\textsuperscript{\ensuremath{\pm}2.18} & 82.48\textsuperscript{\ensuremath{\pm}1.49} & \textbf{23.27}\textsuperscript{\ensuremath{\pm}0.37} \\
    Stair carry & 144.35\textsuperscript{\ensuremath{\pm}0.66} & 45.88\textsuperscript{\ensuremath{\pm}0.22} & 109.96\textsuperscript{\ensuremath{\pm}2.37} & 19.40\textsuperscript{\ensuremath{\pm}0.29} & \textbf{14.71}\textsuperscript{\ensuremath{\pm}0.02} \\
    Precise placement & 78.23\textsuperscript{\ensuremath{\pm}0.21} & 96.12\textsuperscript{\ensuremath{\pm}0.29} & 106.35\textsuperscript{\ensuremath{\pm}0.83} & 39.26\textsuperscript{\ensuremath{\pm}0.72} & \textbf{25.61}\textsuperscript{\ensuremath{\pm}0.46} \\
    Box handover & 88.74\textsuperscript{\ensuremath{\pm}0.98} & 59.38\textsuperscript{\ensuremath{\pm}0.05} & 72.03\textsuperscript{\ensuremath{\pm}0.96} & 10.65\textsuperscript{\ensuremath{\pm}0.43} & \textbf{6.03}\textsuperscript{\ensuremath{\pm}0.06} \\
    Palanquin & 56.58\textsuperscript{\ensuremath{\pm}0.13} & 20.58\textsuperscript{\ensuremath{\pm}0.26} & \textbf{15.11}\textsuperscript{\ensuremath{\pm}0.70} & 24.37\textsuperscript{\ensuremath{\pm}0.22} & 23.90\textsuperscript{\ensuremath{\pm}0.13} \\
    Rebar & 50.66\textsuperscript{\ensuremath{\pm}0.19} & 18.64\textsuperscript{\ensuremath{\pm}0.61} & 70.52\textsuperscript{\ensuremath{\pm}2.57} & 15.18\textsuperscript{\ensuremath{\pm}0.80} & \textbf{14.90}\textsuperscript{\ensuremath{\pm}0.26} \\
    Fire hose & 134.36\textsuperscript{\ensuremath{\pm}0.00} & \textbf{18.62}\textsuperscript{\ensuremath{\pm}0.52} & 138.08\textsuperscript{\ensuremath{\pm}0.56} & 23.10\textsuperscript{\ensuremath{\pm}0.11} & 27.64\textsuperscript{\ensuremath{\pm}0.08} \\
    Box push & 31.12\textsuperscript{\ensuremath{\pm}0.16} & 7.83\textsuperscript{\ensuremath{\pm}0.03} & \textbf{6.68}\textsuperscript{\ensuremath{\pm}0.04} & 13.67\textsuperscript{\ensuremath{\pm}0.01} & 7.32\textsuperscript{\ensuremath{\pm}0.01} \\
    Millstone & \textbf{\ensuremath{<}0.01}\textsuperscript{\ensuremath{\pm}0.00} & \ensuremath{<}0.01\textsuperscript{\ensuremath{\pm}0.00} & \ensuremath{<}0.01\textsuperscript{\ensuremath{\pm}0.00} & \ensuremath{<}0.01\textsuperscript{\ensuremath{\pm}0.00} & \ensuremath{<}0.01\textsuperscript{\ensuremath{\pm}0.00} \\
    \bottomrule
  \end{tabular*}
  \caption{Per-task comparison of the low-level controllers in simulation.
  Each entry reports the mean over three seeds with 512 teams per seed.
  Superscripts show the sample standard deviation across seeds.
  Success rate is the percentage of teams completing a task.
  Body and object errors are mean position errors in centimeters over the full
  reference horizon, including frames after a failure.}
  \label{tab:supp_per_task_ablation}
\end{table*}

\section{Supplementary Videos}
\label{sec:supp_videos}
All videos are on the \projectpage.
The webpage contains a summary video; the spoken team commands with sound, with either a robot or a person as the leader; videos of all real-world experiments in \cref{tab:real}; and videos of the tasks in \bench in simulation, including a long-horizon task composed of several of them.

\end{document}